\documentclass[letterpaper, 10 pt, journal, twoside]{IEEEtran}

\IEEEoverridecommandlockouts                              
                                                          
\usepackage{amsmath,mathtools}
\usepackage{graphics} 
\usepackage{amssymb}  
\usepackage[bookmarks=true]{hyperref}
\usepackage{xcolor,colortbl}
\usepackage{color}
\usepackage{siunitx}
\usepackage{booktabs}
\usepackage{bm}
\usepackage[vlined,ruled,linesnumbered]{algorithm2e}
\usepackage{tikz}
\usetikzlibrary{svg.path}
\usepackage{titlesec}

\usepackage[utf8]{inputenc}
\usepackage{pgfplots}
\DeclareUnicodeCharacter{2212}{−}
\usepgfplotslibrary{groupplots,dateplot}
\usetikzlibrary{patterns,shapes.arrows}
\pgfplotsset{compat=newest}

\newcommand{\algrule}[1][.4pt]{\par\vskip.2\baselineskip\hrule height #1\par\vskip.2\baselineskip}

\newcommand{\hide}[1]{}
\DeclareMathOperator*{\argmin}{argmin}

\newcommand{\rom}[1]{(\expandafter{\romannumeral #1\relax})}
\newcommand{\mat}[1]{\begin{bmatrix}#1\end{bmatrix}}

\let\vec\bm
\newcommand{\lb}[1]{\underbar{$#1$}}
\newcommand{\ub}[1]{\overline{#1}}

\newcommand{\rebuttal}[1]{{#1}}
\newcommand{\rebuttall}[1]{{#1}}

\definecolor{Gray}{gray}{0.9}
\definecolor{somegray}{rgb}{0.5, 0.5, 0.5}

\makeatother

\titlespacing*{\section}
{0pt}{1.0ex plus .2ex minus .2ex}{0.2ex minus .2ex}
\titlespacing*{\subsection}
{0pt}{1.0ex plus .2ex minus .2ex}{0.2ex minus .2ex}
\titlespacing*{\subsubsection}
{0pt}{0.0ex plus .2ex minus .2ex}{0.0ex minus .2ex}
\titlespacing*{\paragraph}
{0pt}{0.0ex plus .2ex minus .2ex}{0.0ex minus .2ex}
\titlespacing*{\subparagraph}
{0pt}{0.0ex plus .2ex minus .2ex}{0.0ex minus .2ex}

\begin{document}

\setlength{\abovedisplayskip}{6pt}
\setlength{\belowdisplayskip}{6pt}
\setlength{\abovedisplayshortskip}{4pt}
\setlength{\belowdisplayshortskip}{4pt}


\title{Time-Optimal Online Replanning for \\Agile Quadrotor Flight}
\author{Angel Romero, Robert Penicka, \rebuttal{and} Davide Scaramuzza
\thanks{\rebuttall{
 Manuscript received: February, 24, 2022; Revised April, 19, 2022; Accepted June, 7, 2022.
 This paper was recommended for publication by Editor T. Ogata upon evaluation of the Associate Editor and Reviewers' comments. This work was supported by the National Centre of Competence in Research (NCCR) Robotics through the Swiss National Science Foundation (SNSF), and the European Union’s Horizon 2020 Research and Innovation Programme under grant agreement No. 871479 (AERIAL-CORE) and the European Research Council (ERC) under grant agreement No. 864042 (AGILEFLIGHT).}}
 \thanks{\rebuttall{The authors are with the Robotics and Perception Group, University of Zurich, Switzerland} (\protect\url{http://rpg.ifi.uzh.ch}). (email: \protect\url{roagui@ifi.uzh.ch})}
\thanks{Digital Object Identifier (DOI): see top of this page.}
}

\maketitle

\begin{abstract}
In this paper, we tackle the problem of flying a quadrotor using time-optimal control policies that can be replanned online when the environment changes or when encountering unknown disturbances.
%
This problem is challenging as the time-optimal trajectories that consider the full quadrotor dynamics are computationally expensive to generate, on the order of minutes or even hours.
%
%
We introduce a sampling-based method for efficient generation of time-optimal paths of a point-mass model.
These paths are then tracked using a Model Predictive Contouring Control approach that considers the full quadrotor dynamics and the single rotor thrust limits.
%
%
Our combined approach is able to run in real-time, being the first time-optimal method that is able to adapt to changes \emph{on-the-fly}.
We showcase our approach's adaption capabilities by flying a quadrotor at more than 60 km/h in a racing track where gates are moving.
Additionally, we show that our online replanning approach can cope with strong disturbances caused by winds of up to 68 km/h.
\end{abstract}

\rebuttall{
}
\rebuttall{
\section*{Supplementary Material}
\rebuttal{\noindent Video of the experiments: \url{https://youtu.be/zBVpx3bgI6E}}
}
\IEEEpeerreviewmaketitle

\section{Introduction}

\IEEEPARstart{Q}{uadrotors} are one of the most versatile flying robots. 
Depending on the task at hand, their design can vary from very robust to extremely fast and powerful agile machines.
When it comes to speed, racing quadrotors are among the fastest and most agile flying devices ever built~\cite{spectrum2020acrobatics} .
Winning a drone race requires a set of visual, coordination and motor skills that are only achieved by the most dexterous and experienced human pilots~\cite{Pfeiffer21ral}.
For this reason, finding algorithms that conquer the sport of drone racing would represent a giant step forward for the entire robotics community.
Thus, drone racing research has not only been feeding from the latest advances in perception, learning, planning, and control, but it has also been a strong contributor to these fields \cite{alphapilot, feigao_fast_racing, CPC, mpcc, beauty_beast, deep_drone_racing, PAMPC, karaman_blackbox}.
A backbone piece in the drone racing is time-optimal (i.e., minimum-time) quadrotor flight through a series of waypoints (i.e., gates).
An extremely challenging and currently missing piece of the puzzle is solving this problem \emph{on-the-fly}.

Planning a minimum-time trajectory is a complex problem.
%
Algorithms need to both come up with the exact sequence of positions, velocities, orientations, inputs, etc. of the platform at every potential situation, and their allocation in time.
Additionally, for this sequence to be tracked by classical control methods, it needs to be dynamically feasible, i.e., it needs to fulfill the complex non-linear dynamics of the quadrotor platform, and its actuator constraints.
%

\begin{figure}[t]
    \centering
    \includegraphics[width=\linewidth]{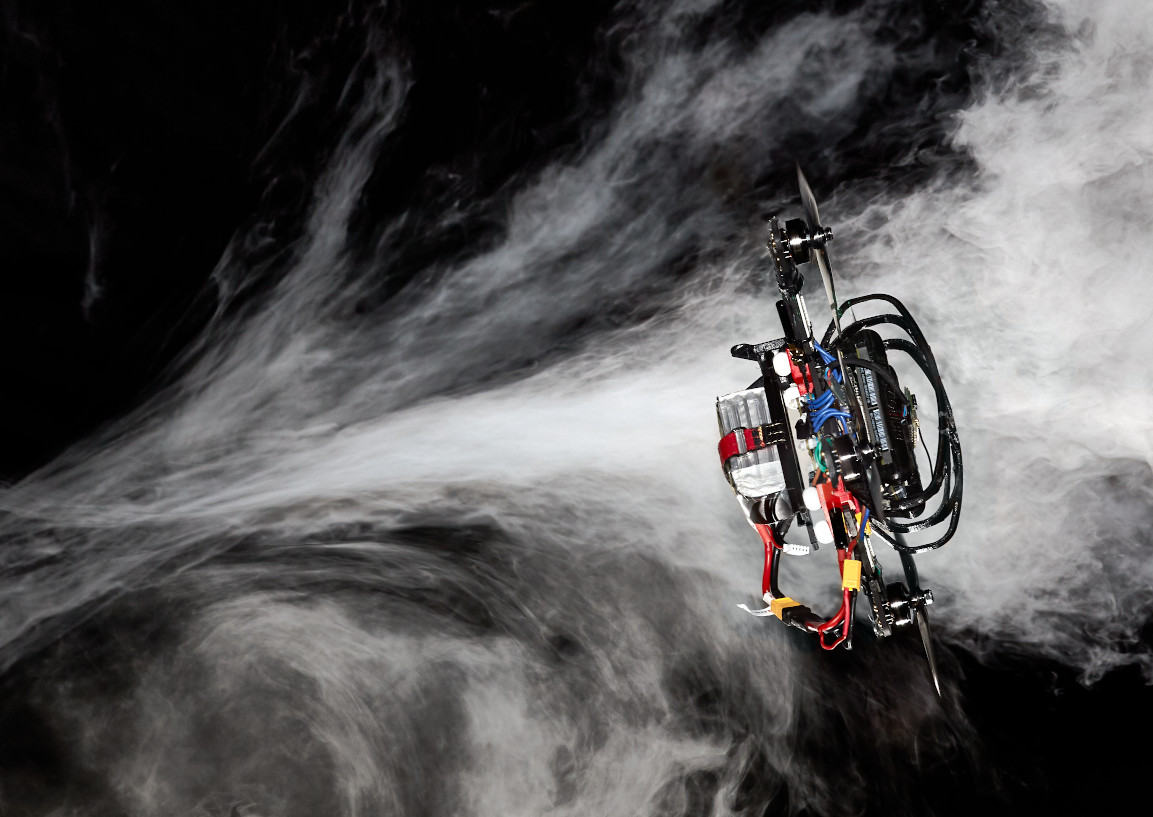}
    \caption{The proposed algorithm is able to adapt \emph{on-the-fly} when encountering unknown disturbances. In the figure we show a quadrotor platform flying at speeds of more than 60 km/h. Thanks to our online replanning method, the drone can adapt to wind disturbances of up to 68 km/h while flying as fast as possible.
    \vspace{-0.7cm}
    }
    \label{fig:eyecatcher}
\end{figure}

Different approaches have attempted to tackle this problem.
In~\cite{CPC}, the authors propose the first method that finds time-optimal trajectories for a given series of waypoints and show how they beat professional human pilots in a drone race. 
%
%
However, these trajectories take several hours to compute, and hence their use is prohibitive for any online adaption to unpredicted changes (e.g., moving gates or wind gusts).
%
To overcome this limitation, we need an algorithm to generate time-optimal trajectories in real-time.
In~\cite{mpcc}, a Model Predictive Contouring Control (MPCC) method is introduced that is able to track non-feasible paths in a near-time-optimal fashion.
By solving the difficult time-allocation problem online, at every time step, the MPCC method optimally selects the states and inputs that maximize the progress along the given path, while minimizing the Euclidean distance to it.
The authors \rebuttal{compare to \cite{CPC} and} show that the resulting policies are very close to time-optimality, opening the possibility of using as reference simplified paths that are faster to generate.

Our contributions are as follows.
We introduce an efficient sampling-based method that can plan time-optimal paths for point-mass model in real-time. 
Additionally, we extend the implementation of Model Predictive Contouring Control~\cite{mpcc} (currently the state of the art in 
time-optimal flight) to make it suitable for online replanning.
%
Our approach exploits the ability of MPCC to track non-feasible paths and uses efficiently-generated, time-optimal point-mass-model \rebuttal{(PMM)} trajectories to enable online replanning, being the first minimum time control method that is able to adapt to changes \emph{on-the-fly}.
We show in real world experiments that our approach is able to fly through a series of gates in minimum time, while the gates are continuously moved.
Additionally, we show that the online replanning ability offers extra robustness to unknown, external disturbances by optimally flying through a strong stream of wind blowing up to 68 km/h.

\section{Related Work}
Our approach combines two well known concepts in quadrotor research: online trajectory replanning and time-optimal flight.
\subsection{Quadrotor online trajectory replanning}

The existing approaches for quadrotor trajectory planning can be mostly categorized as polynomial approaches, sampling-based approaches, and optimization methods. 

The idea of representing trajectories as continous time polynomials has been a recurrent topic of reserach in quadrotor planning \cite{Mueller15cdc, Richter13isrr, feigao_fast_racing}. 
This is possible due to the fact that quadrotors are differentially flat systems \cite{Mellinger12ijrr}, which allows for sampling the quadcopter inputs from independent polynomials in the $x$, $y$, $z$ and yaw axes. 
Polynomials are very computationally efficient and thus allow online replanning.
However, polynomial trajectory representation cannot render time-optimal policies, since the polynomials are smooth by definition.

Other methods that also take advantage of the differential flatness property are \cite{feigao_raptor, feigao_gradient, feigao_trajectory_generation}. 
These methods use polynomial interpolations (splines) for trajectory optimization, and therefore, are also efficient enough for online replanning. 
However, even if they pursue fast flight, they jointly optimize perception objectives, smoothness, dynamic feasibility, collisions, etc., making their performance far from being minimum time.

\subsection{Time-optimal quadrotor flight}
Time optimal quadrotor flight has been  a very active area of research in the last years.
The oldest approaches that have worked on this direction are \cite{Loock13ecc, Hehn12ar}, but they have strong limitations. Either they reduce the problem to 2 dimensions, or make assumptions under which the time-optimal input policy is a pure bang-bang acceleration.
More recent approaches \cite{Spedicato18tcst} use the arc length to frame the problem independently of the time variable. However, they also suffer from limitations, since they simplify the limits of the platform to collective thrust and body rates, instead of single rotor thrusts.

The first approach that takes into account the full quadrotor dynamics to plan a pure time-optimal trajectory through a series of waypoints is \cite{CPC}.
They consider the time-optimal trajectory generation problem in a discretized way and use optimization-based methods to assign a time allocation and a full state to each trajectory point.
This method has shown to result in very good performance, however, it suffers from a major drawback.
This algorithm takes a considerably long amount of time to converge (from 30 minutes to several hours, depending on the track length). 
This cripples its ability to adapt to changes or to model mismatches or unknown disturbances on the fly, bringing up two limitations.
The first being the need for a control safety margin.
The second being its inability to adapt to changes in the track or to major disturbances.
In \cite{mpcc}, the first limitation is tackled. 
Since the time allocation problem is solved online, this approach can optimally select the states and inputs in the prediction horizon at every timestep in order to maximize progress along the track, and is therefore able to make full use of the quadrotor's actuation power even in the presence of model mismatch.
Another major benefit of \cite{mpcc} is that it only needs a continuously differentiable 3D path to track, and not a fully feasible trajectory. This allows for the use of simplified paths as reference, such as a time-optimal point mass model trajectory.


\section{Methodology}
\subsection{Quadrotor Dynamics}
\label{sec:quad_dynamics}
The quadrotor's state space is described from the inertial frame $I$ to the body frame $B$, as $\vec{x} = [\vec{p}_{IB}, \vec{q}_{IB}, \vec{v}_{IB}, \vec{w}_{B}]^T$ where $\vec{p}_{IB}\in \mathbb{R}^3$ is the position, $\vec{q}_{IB} \in \mathbb{SO}(3)$ is the unit quaternion that describes the rotation of the platform, $\vec{v}_{IB} \in \mathbb{R}^3$ is the linear velocity vector, and $\vec{\omega}_{B} \in \mathbb{R}^3$ are the bodyrates in the body frame.
The input of the system is given as the collective thrust $\vec{f}_B = \mat{0 & 0 & f_{Bz}}^T$ and body torques $\vec{\tau}_B$.
For readability, we drop the frame indices as they are consistent throughout the description.
The dynamic equations are
\vspace{-0.25cm}
\begin{gather}
\begin{aligned}
\dot{\vec{p}} &= \vec{v} & \quad
\dot{\vec{q}} &= \frac{1}{2} \vec{q} \odot \mat{0 \\ \vec{\omega}} \\
\dot{\vec{v}} &= \vec{g} + \frac{1}{m} \mathbf{R}(\vec{q}) \vec{f}_T &
\dot{\vec{\omega}} &= \mathbf{J}^{-1} \left( \vec{\tau} - \vec{\omega} \times \mathbf{J} \vec{\omega} \right)
\label{eq:quad_dynamics}
\end{aligned}
\end{gather}
where $\odot$ represents the Hamilton quaternion multiplication, $\mathbf{R}(\vec{q})$ the quaternion rotation, $m$ the quadrotor's mass, and $\mathbf{J}$ the quadrotor's inertia.

Additionally, the input space given by $\vec{f}$ and $\vec{\tau}$ is decomposed into single rotor thrusts $\vec{f} = [f_1, f_2, f_3, f_4]$ where $f_i$ is the thrust at rotor $i \in \{ 1, 2, 3, 4 \}$.
\begin{align}
\vec{f}_T &= \mat{0 \\ 0 \\ \sum f_i} &
\text{and }
\vec{\tau} &=
\mat{l/\sqrt{2} (f_1 + f_2 - f_3 - f_4) \\
l/\sqrt{2} (- f_1 + f_2 + f_3 - f_4) \\
c_\tau (f_1 - f_2 + f_3 - f_4)}
\label{eq:quad_inputs}
\end{align}
with the quadrotor's arm length $l$ and the rotor's torque constant $c_\tau$.

Furthermore, in order to approximate the most prominent aerodynamic effects, we extend the quadrotor's dynamics to include a linear drag model \cite{Faessler18ral}. Let $\bm{D}$ be the diagonal matrix with drag coefficients $d$ such that $\bm{D} = \text{diag}\left(d_x, d_y, d_z\right)$, we expand the dynamic model from \rebuttal{\eqref{eq:quad_dynamics}} by:
\begin{equation}
\dot{\bm{v}} = \bm{g} + \frac{1}{m} \bm{R}(\bm{q}) \bm{f_T} - \bm{R}(\bm{q}) \bm{D} \bm{R}^\intercal(\bm{q}) \cdot \bm{v}
\end{equation}

\subsection{Model Predictive Contouring Control}
Model Predictive Contouring Control (MPCC) \cite{mpcc} offers a change of paradigm with respect to classical Model Predictive Control approaches.
Instead of minimizing the error in the input and state space, with respect to a given trajectory, it minimizes the euclidean distance to a given 3D path, while maximizing how fast the path is travelled.
One of the benefits of this control approach is that it does not need a feasible trajectory to be generated beforehand.
It is sufficient to provide the controller with a continuously differentiable 3D path.
It achieves this by solving the difficult time allocation problem and the control problem concurrently, and in real time.
The main cost function to be solved by MPCC is the following:
\vspace{-0.2cm}
\begin{equation} \label{eq:full_ocp}
\begin{split}
  \pi(\vec{x}) =  \argmin_u 
   & \sum_{k = 0}^{N } \Vert \vec{e}^l(\theta_k) \Vert_{q_l}^2 + \Vert \vec{e}^c(\theta_k) \Vert_{q_c}^2 \\  
    + \Vert\vec{\omega}_k\Vert_{\bm{Q_{\omega}}}^2 
    &+ \Vert\Delta v_{{\theta}_k}\Vert_{r_{\Delta v}}^2 + \Vert\Delta\vec{f}_k\Vert_{\bm{R_{\Delta f}}}^2 - \mu v_{\theta,k} \\
  \text{subject to }&  \vec{x}_0 = \vec{x}  \\
  &\vec{x}_{k+1} = f(\vec{x}_k, \vec{u}_k)  \\
  &\lb{\vec{\omega}} \leq \vec{\omega} \leq \ub{\vec{\omega}}  \\
  &\lb{\vec{f}} \leq \vec{f} \leq \ub{\vec{f}}  \\
  & 0 \leq v_{\theta} \leq \ub{v_{\theta}}  \\
  & \lb{\Delta v}_{\theta} \leq \Delta v_{\theta} \leq \ub{\Delta v_{\theta}}  \\
  & \lb{\Delta \vec{f}} \leq \Delta \vec{f} \leq \ub{\Delta \vec{f}}
\end{split}
\end{equation}

where $f(\vec{x}_k, \vec{u}_k)$ is the dynamic equation, described in Section \ref{sec:quad_dynamics}, $\theta_k$ is the so-called \emph{progress} term at timestep $k$, and \rebuttal{$v_{\theta, k}$}, it's derivative with respect to time. $\vec{e}^c(\theta_k)$ is the \emph{contour error}, which is defined as the minimum euclidean distance from the current position to the reference 3D path, \rebuttal{and $\vec{e}^l(\theta_k)$ is the \emph{lag error}. These errors are penalized by the weights $q_c$ and $q_l$, respectively. $q_c$ is dynamically changed depending on the position of the waypoints to accurately pass through them. For more details on this control approach, we refer the reader to \cite{mpcc}.}

\subsection{Time-Optimal Point-Mass Model Trajectory}\label{sec:offline_traj_optimization}
\label{sec:PMM}
As mentioned in the previous section, one of the benefits of MPCC is that its only requirement is a continuously differentiable 3D path that does not need to be dynamically feasible.
In \cite{mpcc}, the authors leverage this advantage and generate paths using a sampling based approach that considers a point-mass model instead of the full quadrotor dynamics for the path generation.
They show that the platform is able to fly in a time-optimal \rebuttal{fashion}, while significantly reducing the computational burden, since the generation of point-to-point trajectories using a point mass model has a closed-form solution.
In what follows, we explain how these time-optimal point-mass model trajectories are generated, and how we have made the method more efficient in order to achieve online replanning capabilities.

Given an initial state consisting of position $\vec{p}_0$ and velocity $\vec{v}_0$, and given the reduced dynamics of a point-mass model, $\ddot{\vec{p}} = \vec{u}$, with the input acceleration being constrained $\lb{\vec{u}} \leq \vec{u} \leq \ub{\vec{u}}$, it can be shown by using Pontryagin's maximum principle \cite{bertsekasDP} that the time-optimal control input results in a bang-bang policy of the form:
\begin{align}
  u_{}^*(t) = \left\{
  \begin{matrix}
    \lb{u}_{}, & 0 \leq t \leq t_1^* & \\
    \ub{u}_{},  & t_1^* \leq t \leq T^* & (= t_1^* + t_2^*)
  \end{matrix}\right.
  \label{eq:time_optimal_x}
\end{align}
or vice versa, starting with $\ub{u}_{x}$. 
The input policy shown in \rebuttal{\eqref{eq:time_optimal_x}} results in the following trajectory:
\begin{equation} \label{eq:motion}
\begin{split}
  p_{1} &= p_{0} + v_{0}t_1 + \frac{\lb{u}_{}}{2} t_1^2 \\
  v_{1} &= v_{0} + \lb{u}_{}t_{1} \\
  p_{2} &= p_{1} + v_{1}t_2 + \frac{\ub{u}_{}}{2} t_2^2 \\
  v_{2} &= v_{1} + \ub{u}_{}t_{2}
\end{split}
\end{equation}
where it is straightforward to verify that, if we know $p_{0}$, $v_{0}$, $p_{2}$, $v_{2}$, $\lb{u}$ and $\ub{u}$, then there are exactly four unknowns: $t_1$, $t_2$, $p_{1}$ and $v_{1}$, and therefore, we can solve \rebuttal{\eqref{eq:motion}} in closed form.
The previous \rebuttal{\eqref{eq:time_optimal_x} and \eqref{eq:motion},} depict the solution for one of the axis. We therefore do this for every axis ($x$, $y$ and $z$) separately.
\vspace{-0.4cm}
\begin{figure}[!htb]
    \centering
    \tikzstyle{m}= [circle,fill,inner sep=1.5pt]
\tikzstyle{n}= [circle,inner sep=-0.7pt]
\newcounter{x}\newcounter{y}
\newcommand{\interval}[2]{$v^{#2}_{#1}$}
\newcommand{\nodes}[2]{
   \node[n] (H#2) at (#1,1.2) {$p_{{#2}}$};
	\node[m,label={\interval{#2}{1}}] (N-#2-1) at (#1,-1) {};
    \node[m,label={\interval{#2}{2}}] (N-#2-2) at (#1,-3) {};
    \node[n] (D#2) at (#1,-3.8) {$\vdots$};
    \node[m,label={\interval{#2}{h}}] (N-#2-3) at (#1,-6) {};
 }
\begin{tikzpicture}[yscale=0.5, xscale=1.1, node distance=0.3cm, auto,
   source/.style={draw,thick,rounded corners,inner sep=.3cm,text width=0.6em, text height=8.5em}]
    \node[m,label={$p_0,v_{0}$}] (init) at (-1.75,-3) {};
    \node[source] (s1) at (0,-3) {};
    \node[source] (s2) at (1.75,-3) {};
    \node[source] (s3) at (3.5,-3) {};
    \nodes{0}{1}
    \nodes{1.75}{2}
    \nodes{3.5}{3}
    
    \path (init) edge[->] (N-1-1);
    \path (init) edge[->] (N-1-2);
    \path (init) edge[->] (N-1-3);
    \foreach \y in {1,...,3}
        \foreach \x in {1,...,3}
        {
            \path (N-1-\x) edge[->] (N-2-\y);
            \path (N-2-\x) edge[->] (N-3-\y);
       }
\end{tikzpicture}
    \caption{Velocity search graph where every column $p_i, i \in [1,2,3]$ is a gate in the receding horizon. We find the minimum time, point-mass model trajectory by doing Dijkstra's algorithm on this graph. The different ways of sampling the nodes on the graph characterize the different algorithms described in this paper.
    \vspace{-0.3cm}
    }
    \label{fig:velocity_search}
\end{figure}
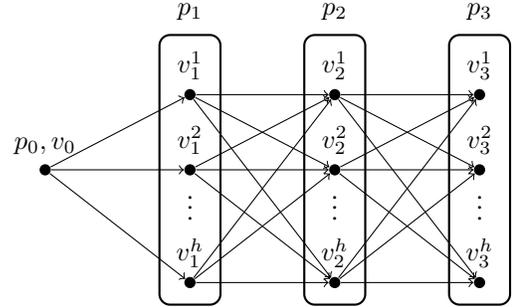

Once we compute the per-axis minimum times ($T_{axis}$), we choose the maximum of these minimum times \mbox{($T^* = \text{max}(T^*_x, T^*_y, T^*_z)$)} and slow down the other two axes' policies to make all three axes' durations equal.
To do this, a new parameter $\alpha \in [0, 1]$ is introduced that scales the acceleration bounds.
For example, if we need to increase the duration of the $x$ axis, the applied control inputs are scaled to $\alpha \lb{u}_x$ and $\alpha \ub{u}_x$, respectively.
In order to find the value of $\alpha$, we solve again \rebuttal{\eqref{eq:motion}} but now with $\alpha$ multiplying the respective accelerations, and we augment \rebuttal{\eqref{eq:motion}} with
\begin{align}
    t_{1, axis} + t_{2, axis} = T^* 
\end{align}
and solve again the augmented system.

Assuming that the positions of the waypoints are known, the idea is to recursively sample the velocities at the waypoints and find the combination of velocities that make the final trajectory time optimal for a point-mass model. 
To solve the problem efficiently, we use a receding horizon approach where only the next $H_g$ gates are taken into account and then treat this as a shortest path problem and solve it using Dijkstra's algorithm \cite{bertsekasDP}.

This is depicted in Fig. \ref{fig:velocity_search}, where every column represents a different waypoint position and for every column, a set of velocities are sampled. There are different ways in which these velocities can be sampled, and they are discussed in the next section.
In our particular application, the gate horizon has been chosen to be $H_g = 3$ gates.  To justify this choice, in \cite{mpcc} the authors have generated PMM trajectories for the task of completing 3 laps on the \emph{CPC-track}, for different values of $H_g$. 
The resulting total lap times are then averaged over 10 runs and compared.
In Table II of \cite{mpcc}, one can notice how for $H_g$ larger than 3, the improvement in total time is negligible.

\subsection{Velocity search strategies}
From Fig. \ref{fig:velocity_search} one can see how the amount of nodes inside the velocity sampling space is the main contributing factor to the amount of edges in our graph, and, therefore, to the amount of times that we need to solve \rebuttal{\eqref{eq:motion}.}
More generally, if there are $h$ velocity samples per gate, and $H_g$ gates, the total number of edges in our velocity graph $E$ is:
\begin{align}
E = h + (h^2 \cdot (H_g - 1))
\end{align}
which means that the upper bound for computational complexity grows quadratically with the number of velocity samples per gate $h$.
This is an upper bound because we use Dijkstra's algorithm \cite{bertsekasDP}, which means that potentially not all edges need to be computed.

Two different velocity search strategies are explored in this work:

\subsubsection{Random sampling}
We randomly sample $h$ different velocities such that they lie in a cone pointing towards the exit direction of the gate. The sampling distribution is uniform, and $h = 150$ has shown to be a good value to achieve time optimal performance \cite{mpcc, alphapilot}. This is the approach that is used for the generation of the fixed PMM reference in our experiments.

The worst case number of edges that we would need to compute for this approach is:
\begin{align}
    E_{random} =& h + (h^2 \cdot (H_g - 1)) \notag\\
    =& 150 + (150^2 \cdot 2) = 45150
    \label{eq:random_n_samples}
\end{align}

\subsubsection{Cone Refocusing via Binary Search}
\begin{figure}[t]
    \centering
    \includegraphics[width=0.9\linewidth]{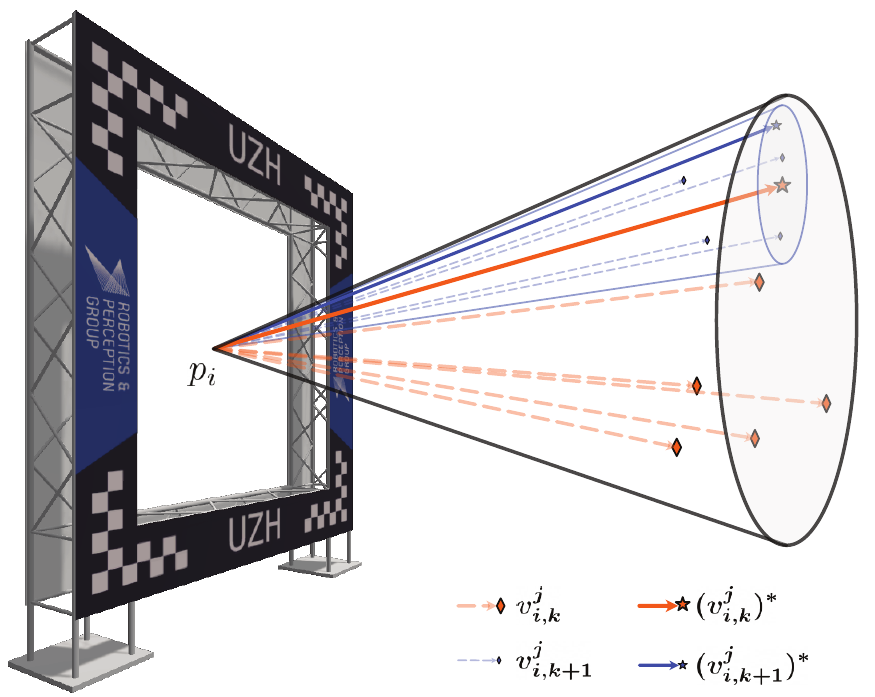}
     \vspace{-0.4cm}
    \caption{One iteration of the cone refocusing recursion. The dashed orange velocity samples are equally distributed over the larger cone. The solid orange arrow represents the optimum over the large cone. Our algorithm refocuses a smaller, second cone around this optimum, and repeats the approach in the smaller cone (blue samples).
    }
    \label{fig:gate_cone_refocus}
\end{figure}

\begin{figure}[t]
    \centering
    \includegraphics{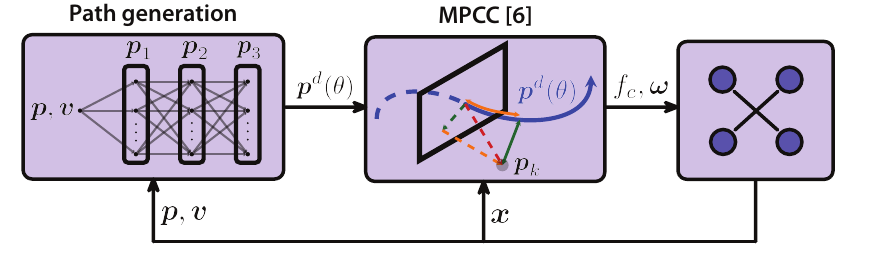}
    \vspace{-0.9cm}
    \caption{\rebuttall{Block diagram of the proposed online replanning approach. At every timestep we generate a new path that considers the current position and velocity and that is tracked by the MPCC controller architecture.\vspace{-0.05cm}}}
    \label{fig:block_diagram}
\end{figure}

In order to reduce the computational burden of the method described above, instead of randomly sampling over a cone in front of the exit direction of the gate, we do it in a more directed way.
First, we bin the cone in front of the gate in the following way: we sample  $s$ different velocity norms, $s$ different values for yaw and $s$ different values for pitch. These values are evenly distributed over the entire feasible velocity space.
This means that in this case, instead of 150, $h = s^3$. In our case, we have chosen $s = 3$, which reduces significantly the number of samples per gate to $h = 27$.
With these samples, we build the nodes of the velocity graph as shown in Fig. \ref{fig:velocity_search}.
Next, we compute the edges of this graph by using \rebuttal{\eqref{eq:motion}, \eqref{eq:time_optimal_x}, and use Djikstra's algorithm to find the optimal path.}
Then, we repeat the process again, but this time instead of sampling the velocities on the entire feasible cone, we \emph{refocus} the cone and sample only on a small cone around the previous minimum time trajectory.
We continue this iterative process until the ratio between previous minimum time and new minimum time is less than $\epsilon$.
In our case, this $\epsilon$ has been chosen such that if the improvement is less than 1\% in minimum time between iteration, the algorithm has converged and the iteration ends.
A more detailed description of this algorithm is shown in Algorithm \ref{alg:PMM}. A graphical illustration of one cone refocusing recursion step is depicted in Fig. \ref{fig:gate_cone_refocus}. \rebuttall{Additionally, a block diagram showing how the proposed planning approach is combined with MPCC is depicted in Fig. \ref{fig:block_diagram}}

Finally, the upper bound number of edges that we need to compute is now:
\begin{align}
E_{refocus} =& (h + (h^2 \cdot (H_g - 1))) \cdot K \notag \\
 =& (27 + (27^2 \cdot 2)) \cdot 4 = 5940
 \label{eq:refocus_n_samples}
\end{align}
where $\rebuttall{K}$ is the number of refocusing iterations. Experimentally, most of the times it takes $K = 4$ iterations for this algorithm to converge.
\vspace{-0.4cm}
\SetAlFnt{\footnotesize}
\setlength{\textfloatsep}{0pt}
\SetKwFor{Loop}{loop}{}{end}
\begin{algorithm}
  \KwIn{$p[i], i \in [1, 2, ... , H_g]$, $x_0$ initial state, $s$ number of bins}
   \algrule
   \DontPrintSemicolon
  \For{$k \in [0, 1, ...]$}{
   $V_{graph} \gets [\enskip]$ \\
  \For{$i \in [1, 2, ... , H_g]$}{
    $p_i \gets p[i]$, $j \gets 0$ \\
    $v_{step} \gets (v_{max} - v_{min}) / v_{grid}$ \\
    $\theta_{step} \gets (\theta_{max} - \theta_{min}) / \theta_{grid}$ \\
    $\psi_{step} \gets (\psi_{max} - \psi_{min}) / \psi_{grid}$ \\
    \For{$l \in [0, s-1]$} {
    $v_l \gets v_{min} + l \cdot v_{step}$\\
     \For{$m \in [0, s-1]$} {
     $\theta_m \gets \theta_{min} + m \cdot \theta_{step}$\\
      \For{$n \in [0, s-1]$} {
      $\psi_n \gets \psi_{min} + n \cdot \psi_{step}$ \\
      $V_{graph}[i,j] \gets vec(v_l, \theta_m, \psi_n)$ \\
      $j \gets j + 1$\\
      }
    }
    }
  }
  $((v_{i,k}^j)^*, T^*_k) \gets$ \eqref{eq:time_optimal_x}, \eqref{eq:motion} and Dijkstra's algorithm on $V_{graph}$\\
  Update search cone using $(v_{i,k}^j)^*$:\\
      \enskip \enskip update $v_{max}, v_{min}$\\
      \enskip \enskip update $\theta_{max}, \theta_{min}$\\
      \enskip \enskip update $\psi_{max}, \psi_{min}$\\
  \If{$T^*_k > \epsilon \cdot T^*_{k-1}$}{
    break\\
  }
  }
  \caption{PMM generation via cone refocusing}
  \label{alg:PMM}
\vspace{-0.18cm}
\end{algorithm}
\vspace{-0.5cm}

\subsection{Comparison}

In this section we compare both sampling approaches introduced above. As one can tell by looking at \rebuttal{\eqref{eq:random_n_samples} and \eqref{eq:refocus_n_samples}}, since the worst case computational complexity of Dijkstra's algorithm evolves quadratically with the number of edges of the graph, we already expect the cone refocusing method to be computationally lighter than the random approach. This is indeed the case, as shown in Table \ref{tab:solve_times}.

\begin{table}[t]
\vspace{-0.2cm}
    \centering
    \setlength{\tabcolsep}{3pt}
    \caption{Computation times for different sampling strategies\vspace{-0.3cm}\label{tab:solve_times}}
    \begin{tabular}{c||c | c}
        \toprule
        \rowcolor{Gray}
        Approach & Solve time (Desktop) & Solve time (Jetson TX2)\\
        \midrule
        Random & 29.67 ms & 126.5 ms \\
        \textbf{Refocus (Ours)} & \textbf{3.48 ms} & \textbf{13.3 ms}\\
        \bottomrule
    \end{tabular}
\end{table}

\begin{figure}[t]
    \centering
    \includegraphics{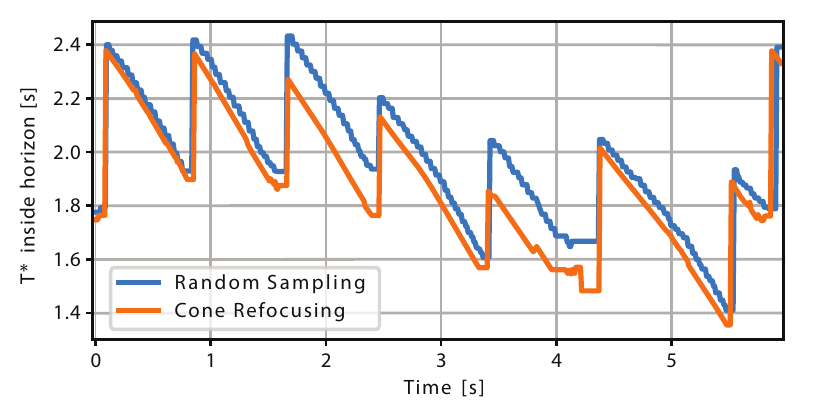}
    \vspace{-0.2cm}
    \caption{Comparison of the optimal time achieved by both sampling approaches over a sequence of gates. The random sampling approach (in blue), limited to a number of samples per gate of 150, always achieves worse or equal minimum times than the proposed cone refocusing approach (in orange).}
    \label{fig:refocusing_comparison}
\end{figure}

However, it still remains the question of which algorithm finds the best solution. In this context, best refers to the one that results in trajectories that achieve a lower $T^*$. To this end, in Fig. \ref{fig:refocusing_comparison} we plot the optimal times for the running horizon, for both approaches.
One can see how the random sampling approach is, at best, equal to the cone refocusing one, and most of the time is worse, since it results in higher minimum times. Additionally, since the random sampling approach cannot run in real-time, it needs to be run every 3-4 control iterations. This is clearly visible in the staircase shape of the blue line in Fig. \ref{fig:refocusing_comparison}. This experiment has been done in a desktop computer.
\rebuttal{The reason for a better performance of the proposed sampling-based method lies in the fact that, in some iterations, e.g., when the distance from the current position to the first gate is very small, the random approach becomes very sensitive to the choice of speed and struggles to find the optimal solution.
Since we have the MPCC controller in the loop, instabilities in the online replanning lead to different MPCC control policies, which can lead to more instabilities, rendering the resulting path sub-optimal (see Fig. \ref{fig:refocusing_comparison}). 

}

\section{Simulation experiments}
\subsection{Experimental Setup}
The identified quadrotor parameters are shown in Table \ref{tab:quads}
\begin{table}[htp]
    \vspace{-0.1cm}
    \centering
    \setlength{\tabcolsep}{3pt}
    \caption{Quadrotor Parameters\vspace{-0.3cm}}
    \label{tab:quads}
    \begin{tabular}{cl||c}
        \toprule
        \rowcolor{Gray}
        \multicolumn{2}{c||}{Parameter} & Nominal Value\\
        \midrule
        $m$& $[\si{\kilo\gram}]$ & $0.752$ \\
        $\rebuttal{\boldsymbol{J}}$& $[\si{\gram\meter^2}]$ & $\mathrm{diag}(2.5, 2.1, 4.3)$  \\
        $\left(u_\mathrm{min}, u_\mathrm{max}\right)$& $[\si{\newton}]$ & $\left(0, 8.5\right)$  \\
        $\left(\rebuttal{d_x,~d_y,~d_z}\right)$& $[\si{kg/s}]$ & $\left(0.26,~0.28,~0.42\right)$ \\
        \bottomrule
    \end{tabular}
    \vspace{-0.2cm}
\end{table}.
For this experiment, we have identified our in-house designed race drone which we then simulated using the state-of-the-art BEM simulator \cite{neurobem}. 
We choose this simulator because it includes very complex aerodynamic effects caused by how the drone moves through air and by the interaction of the propellers with themselves. This simulator has shown excellent simulation to reality transfer. 
In fact, so much so, that throughout this paper all tuning parameters have been kept unchanged between simulation experiments and real world experiments.
The MPCC controller runs at a frequency of $100$ Hz, and the online replanning is done at every time step, when running in a desktop computer, or every 2 control iterations, when running in the NVIDIA Jetson TX2\footnote{\url{https://developer.nvidia.com/EMBEDDED/jetson-tx2}}.
The simulation experiments presented in this section will be focused on benchmarking our re-planning approach in the \emph{SplitS} race course, the same race track that has been used in \cite{mpcc, CPC}.
We will focus our comparison in three different cases: nominal case, where we don't include any waypoint change or external disturbance; moving waypoint, task that cannot be tackled by the offline planning approach; and wind disturbance case, where we create a strong wind disturbance right before passing through one of the waypoints and compare how the different planning strategies help coping with it.

\subsection{Nominal case}


\begin{figure}[t]
    \centering
    \includegraphics{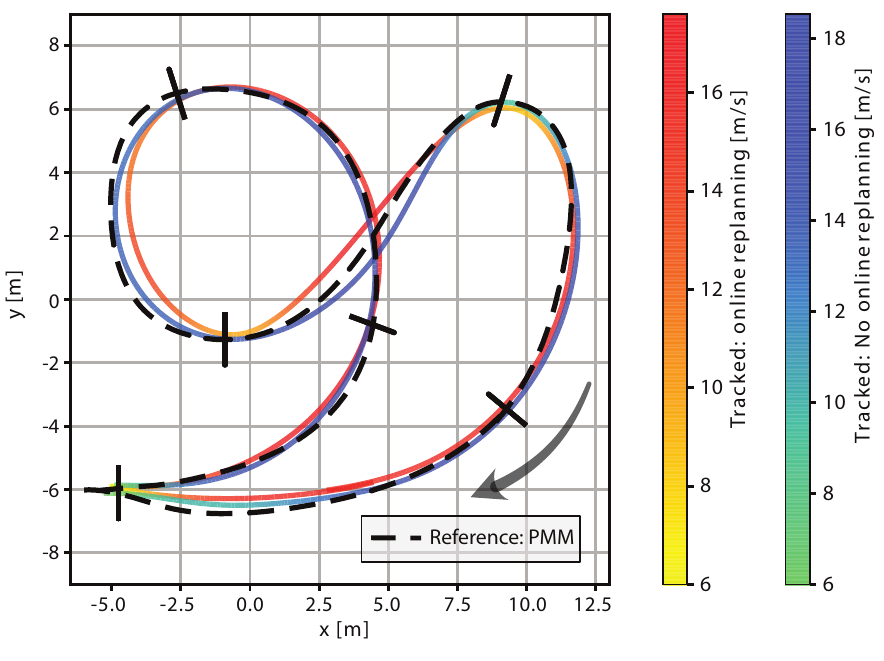}
    \vspace{-0.8cm}    
    \caption{
    Comparison in simulation with and without replanning, in the nominal case without disturbances.}
    \label{fig:sim_nominal}
\end{figure}

The aim of this experiment is to show that even under no external disturbance and under no waypoint movement, the use of replanning still presents an advantage.

In Fig. \ref{fig:sim_nominal}, we show a superposition of our online replanning approach and the one without online replanning, which is tracking the curve labelled as \emph{Reference: PMM}. It is interesting \rebuttal{to} notice that even though both approaches present a very similar shape, our online replanning strategy presents slightly smaller maximum velocities, as one can tell by looking at the colorbars.
This might seem counterintuitive when looking at the measured laptimes, in Table \ref{tab:sim_lap_times}, since the online replanning approach produces lower laptimes.
This is due to the fact that, because we are planning always from the current drone state, the contour error in the first few control horizon steps is very low, enabling for a better maximization of the progress overall. On the contrary, when the reference is fixed, the contour error will generally be non-zero. 
\rebuttal{This can also be seen when measuring the average contour error $C_{e^c} = \frac{\sum_{i = 0}^{M}{\Vert \vec{e}^c(\theta_i)} \Vert^2_{q_c}}{M}$ and average progress velocity $C_{v_\theta} = \frac{\sum_{i = 0}^{M} \mu v_{\theta, i}}{M}$ from the tracks in Fig. \ref{fig:sim_nominal}. For fixed planning, $C_{e_c} = 26120$ and $C_{v_\theta} = 30750$, whereas for replanning, $C_{e_c} = 0$ and $C_{v_\theta} = 33932$.
As predicted, in this case the replanning approach results in higher objectives when maximizing the progress.


}

\subsection{Response to wind disturbance}
\label{sec:wind_disturbance_sim}
In this section we analyze how the online replanning strategy used in this paper helps when reacting to unknown disturbances.
\begin{figure}[t]
    \centering
    \includegraphics{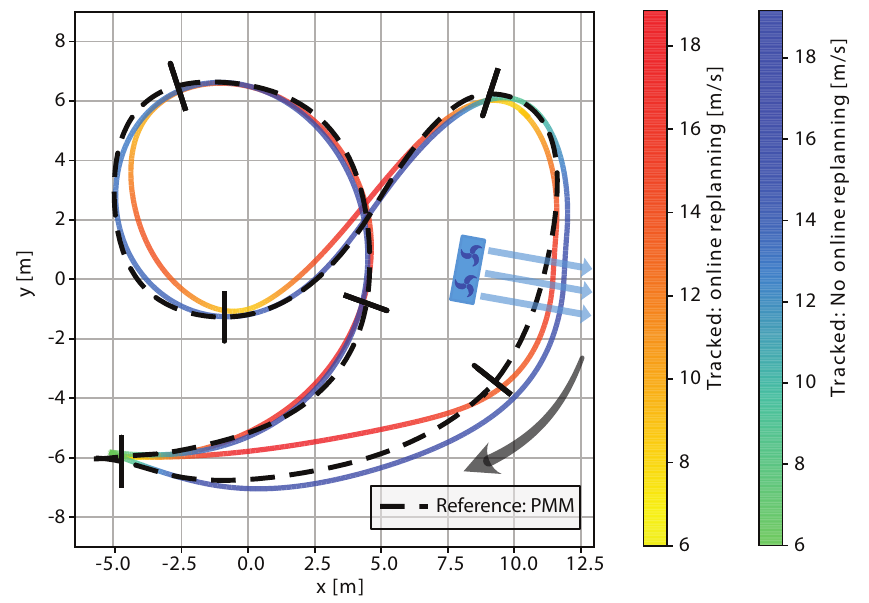}
    \vspace{-0.8cm}
    \caption{Comparison in simulation of response to wind disturbance, with and without replanning. The blue arrows represent the direction of the wind disturbance. Our online replanning approach is able to steer the drone back to the gate.}
    \label{fig:wind_sim}
\end{figure}

Fig. \ref{fig:wind_sim} shows the trajectory tracking performance for the different cases: with and without online replanning. The blue square shows where the wind disturbance has been placed for this experiment, and the arrows show its direction. In the simulator, the wind has been modeled as a constant force of $25$ N in front of the area depicted in blue in Fig. \ref{fig:wind_sim}.
We can see how, given this disturbance, the blue path does not pass through the gate located in the affected area. However, because of the replanning, the proposed strategy is able to consider its current position and velocity in the new plan and successfully pass through the gate.
In Table \ref{tab:sim_lap_times}, we show the minimum lap times of our replanning approach and for the fixed reference approach, for both nominal case and with strong wind disturbances.
\rebuttal{For completeness, a row containing the random sampling approach laptimes has been added. It is important to note that for this case, the planning is not run at every timestep, but every $~30$ ms, as shown in Table \ref{tab:solve_times}.
From these results different details can be highlighted. First, the laptime of the fixed planning approach when presented with wind disturbance is smaller than for our approach. This is because the disturbance pushes the radius of the curve to be larger, and therefore less normal acceleration is needed and higher velocities can be achieved.
However, the lap is invalid because the platform does not pass through the gate.

Second, the laptime of the random sampling approach in the nominal case, is slightly slower than for the fixed planning in the nominal case. 
This is attributed to, as shown in Fig. \ref{fig:refocusing_comparison}, the fact that the random sampling generally generates worse optimal times than the cone refocusing approach. Another reason for this is that the MPCC controller tolerates better small changes in the reference trajectory, and this is only the case when we either replan at every time step or if we don’t replan at all.

Summarizing, even with wind disturbances affecting very close to the waypoint, our approach is able to promptly correct for the external force and pass through the waypoint accurately, while still retaining time optimality.
}

\begin{table}[htp]
    \vspace{-0.2cm}
    \centering
    \setlength{\tabcolsep}{3pt}
    \caption{Simulated minimum lap times\vspace{-0.3cm}}
    \label{tab:sim_lap_times}
    \begin{tabular}{c||c | c}
        \toprule
        \rowcolor{Gray}
        Approach & Nominal case & External disturbance case\\
        \midrule
        Fixed planning \rebuttal{\cite{mpcc}} & 5.67 s & {\color{black} \rebuttal{5.51 s (Invalid)}}\\ 
        \rebuttal{Random Sampling} & \rebuttal{5.74} s & \rebuttal{5.71} s \\
        \textbf{ \rebuttal{Cone refocusing (Ours)}} & \bf{5.57 s} & \bf{5.63 s}\\
        \bottomrule
    \end{tabular}
    \vspace{-0.2cm}    
\end{table}


\section{Real world experiments}
To further display the capabilities of the proposed online replanning method, in this section, we deploy the algorithm on the physical platform.
This platform has been built in-house from off-the-shelf components and it carries an NVIDIA Jetson TX2 board.
\rebuttal{A more detailed description of our platform can be found in \cite{agilicious}.}
Our method can run in both, an offboard desktop computer equipped with an Intel(R) Core(TM) i7-8550 CPU @ 1.80GHz, and on the Jetson TX2.
A Radix FC board that contains the Betaflight\footnote{\url{https://betaflight.com/}} firmware is used as a low level controller. This low-level controller takes as inputs body rates and collective thrusts.
In this case, the body rates are commanded directly from the MPCC controller, and the collective thrust is computed as the sum of the single rotor thrusts.
It is, however, still advantageous that the MPCC considers the full state dynamics. This way, the body rates, and collective thrust are generated taking into account possible current and predicted saturations at the single rotor thrust level.

For state estimation, we use a VICON\footnote{\url{https://www.vicon.com/}} system with 36 cameras in one of the world's largest drone flying arenas ($30\times 30\times 8$\,m) that provide the platform with down to millimeter accuracy measurements of position and orientation. 


\subsection{Implementation Details}
In order to deploy our MPCC controller, \rebuttal{\eqref{eq:full_ocp}} needs to be solved in real-time. 
To this end, we have \rebuttal{implemented our optimization problem using acados \cite{acados} as a code generation tool,} in contrast to \cite{mpcc}, where \rebuttal{its previous version,} ACADO~\cite{ACADO} was used.
\rebuttal{It is important to note that for consistency, the optimization problem that is solved online is written in \eqref{eq:full_ocp} and is exactly the same as in \cite{mpcc}.}
The main benefit of using acados is that it provides an interface to HPIPM (High Performance Interior Point Method) solver~\cite{hpipm}. HPIPM solves optimization problems using BLASFEO~\cite{blasfeo}, a linear algebra library specifically designed for embedded optimization. 
The usage of this new tools and new solver shows a large \rebuttal{decrease in solve times} that allows the deployment of our MPCC algorithm in an embedded platform, in real time.
The solve times and their comparison with the previous implementation are shown in Table \ref{tab:solve_times_acados}.
Our MPCC is run at a $100$ Hz frequency, and the prediction steps are of $60$ ms, with a prediction horizon length of 20 steps.

\begin{table}[htp]
\vspace{-0.3cm}
    \centering
    \setlength{\tabcolsep}{3pt}
    \caption{Solve times MPCC with different implementations\vspace{-0.3cm}}
    \label{tab:solve_times_acados}
    \begin{tabular}{c||c | c}
        \toprule
        \rowcolor{Gray}
        Implementation & Desktop & NVIDIA Jetson TX2\\
        \midrule
        ACADO+QPOASES\footnote{\url{https://github.com/coin-or/qpOASES}}  & 4.8 ms & {\color{black} Doesn't converge}\\
        \textbf{acados+HPIPM (Ours)} & \bf{2.21 ms} & \bf{6.6 ms}\\
        \bottomrule
    \end{tabular}
    \vspace{-0.3cm}
\end{table}

Thanks to this re-implementation, we can run the MPCC together with the online replanning proposed in this paper in the Jetson TX2. 
To achieve this, the planning is running in a different thread, and triggered once every 2 control iterations.

\subsection{Nominal case}
In this section we compare the tracking performance in the \emph{SplitS} race track, in real flight. 
The tracking performance is shown in Fig. \ref{fig:real_world_moving_gate}. Both approaches achieve similar speed of around $18$ m/s, and the behaviour is very similar to the one shown for the same case in the simulation experiments.

\begin{figure}[t]
    \centering
    \includegraphics{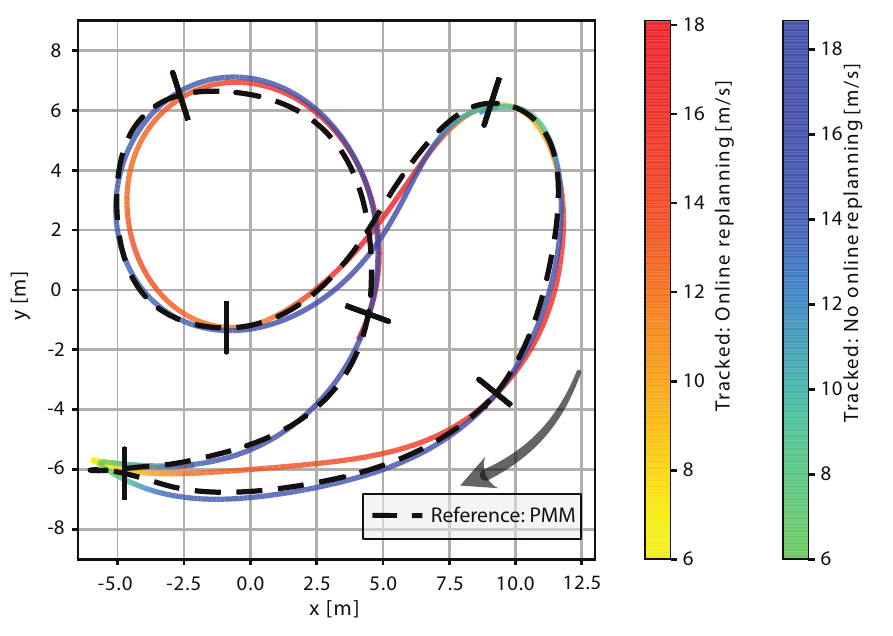}
    \vspace{-0.9cm}
    \caption{Comparison in real flight with and without replanning in the nominal case without disturbances.
    \vspace{-0.05cm}}
    \label{fig:real_world_nominal_case}
\end{figure}

In Table \ref{tab:real_lap_times} we see how the laptimes for the proposed approach are also lower than for the fixed reference case. This follows the same explanation we mentioned in the simulation experiments section: the MPCC controller has less contour error because of the replanning, and this allows for a better maximization of the progress term.

\subsection{Moving waypoint}
\begin{figure}[t]
    \centering
    \begin{tikzpicture}[      
        every node/.style={anchor=south west,inner sep=0pt},
        x=1mm, y=1mm,
      ]   
    \node at (0,0) {\includegraphics[width=\linewidth]{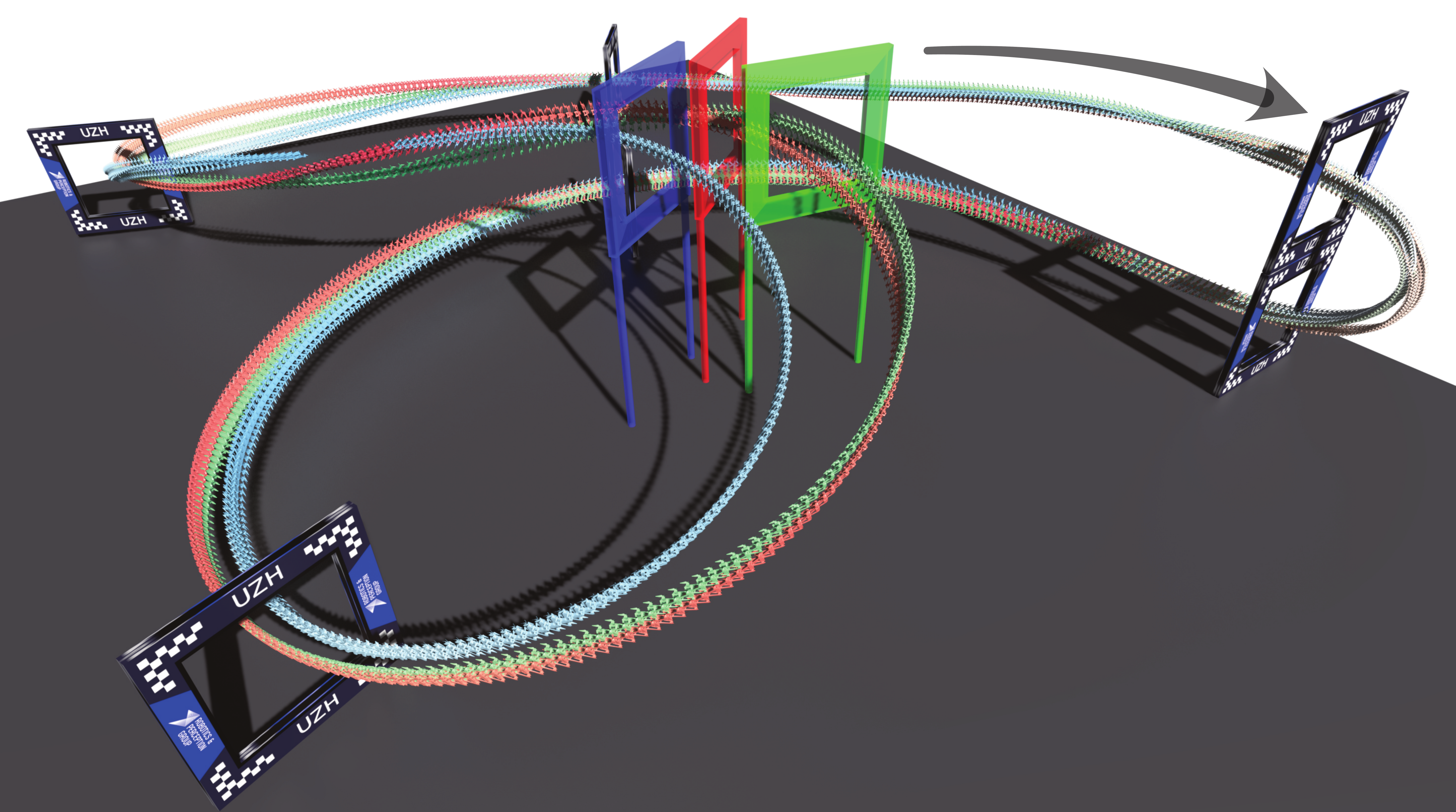}};
    \node at (61,1) {\includegraphics[height=0.22\linewidth]{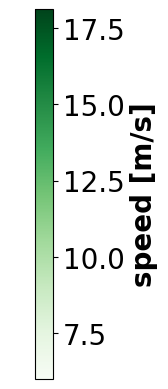}};
    \node at (70,1) {\includegraphics[height=0.22\linewidth]{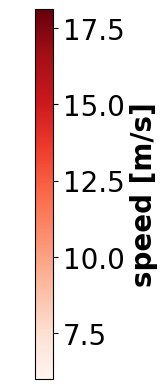}};
    \node at (79,1) {\includegraphics[height=0.22\linewidth]{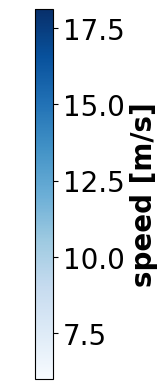}};
    \end{tikzpicture}
    \caption{Real world experiment when waypoints are moving. The red, blue and green gates represent the same gate at different times. The different colored trajectories represent different real-time replanning. Our approach is able to adapt to the movement of the gate while flying as fast as possible.}
    \label{fig:real_world_moving_gate}
\end{figure}

With this experiment we aim to showcase the ability of our approach to steer the platform through the waypoints even in the case when they move. 
Fixed time-optimal reference generating methods cannot adapt to changing conditions, since they cannot run in real-time. Our method can run at every control iteration, therefore, assuming perfect knowledge of the waypoints, we can directly plan a new trajectory that passes through them.
In Fig. \ref{fig:real_world_moving_gate} we show how the drone perfectly flies through all the waypoints, even when the gate that is in the center of the frame is constantly moving.

\subsection{Response to wind disturbance}
In this section we analyze how the online replanning strategy used in this paper helps when reacting to unknown disturbances in the real world. To this end, with the purpose of generating a very strong stream of wind we adapt our experimental setup. 

%
%
%
%

We have placed 4 propellers fixed to a point that is located at $0.5$ m to where our platform flies. These propellers are constantly at full thrust, in order to generate a strong, constant stream of wind with velocities of up to $68$ km/h.

\begin{figure*}[t]
    \centering
    \includegraphics{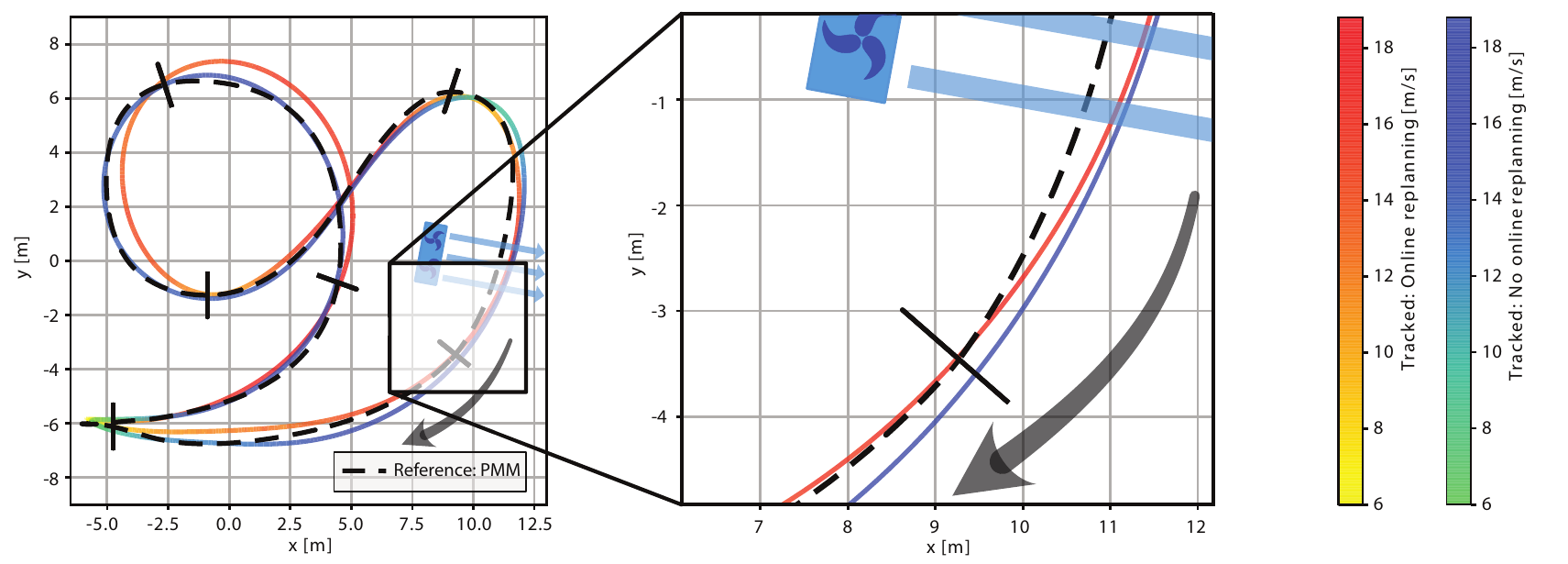}
    \vspace{-0.4cm}
    \caption{Comparison in the real world of response to wind disturbance of 18.8 m/s, with and without replanning. Our online replanning approach (red line) is able to quickly correct for the disturbance and steers the drone back to the center of the gate.
    \vspace{-1.6cm}
    }
    \label{fig:wind_real}
\end{figure*}

In Fig. \ref{fig:wind_real}, we show the tracking performance of the proposed approach compared to the tracking performance of the fixed reference approach.
On the zoomed plot, on the right of Fig. \ref{fig:wind_real}, one can see how the proposed approach flies exactly through the center of the gate, while the fixed reference approach gets deviated from the reference path by $0.24$ m.

\rebuttal{
Similarly to the simulation results, the lap times for the wind disturbance case, shown in Table \ref{tab:real_lap_times}, are only slightly worse than for the non wind disturbance case. Additionally, for the fixed planning approach, in the wind disturbance case the laptimes are smaller than for the nominal case, due to, as explained in \ref{sec:wind_disturbance_sim}, the slightly larger curvature radius.
}
\vspace{-0.3cm}
\begin{table}[!htp]
    \centering
    \setlength{\tabcolsep}{3pt}
    \caption{Real flight minimum lap times\vspace{-0.3cm}}
    \label{tab:real_lap_times}
    \begin{tabular}{c||c | c}
        \toprule
        \rowcolor{Gray}
        Approach & Nominal case & External disturbance case\\
        \midrule
        Fixed planning & 5.93 s & 5.86 s\\
        \textbf{Online replanning (Ours)} & \bf{5.73 s} & \bf{5.83 s}\\
        \bottomrule
    \end{tabular}
\end{table}
\vspace{-0.3cm}

\section{Conclusion}
%


This paper introduces a novel way of coping with dynamic environments online by replanning at every timestep and flying minimum time policies.
We showed that our approach can adapt online to changing position of a gate during drone racing with speeds of more than 60 km/h.
Moreover, we showed that the online replanning strategy can cope with strong wind disturbances. 
However, the state estimate has completely been provided by a motion capture system.
Therefore, a very interesting future direction would be the adaption of the current approach such that it relies only on vision sensors.
For that, perception-awareness of the control approach would be a first necessary step to take, such that the image sensor would always be pointing towards the next waypoint without sacrificing flying performance.
Secondly, vision-based state estimation of both, the platform state and the position of the gates.

\rebuttal{Another connection to our work goes in the direction of sampling-based MPC (SBMPC) \cite{sampling_based_mpc_guidance, sampling_based_mpc_cars}, where also a receding horizon problem is addressed by sampling, but in this case to directly generate feasible control inputs instead a nominal path to track.
We believe it might be an interesting future direction to apply sampling techniques from SBMPC to our planning approach.
}

\rebuttall{
\section{Acknowledgments}
The authors want to thank Thomas Laengle and Cafer Mertcan Akcay for their support with the experimental setup.
}

\bibliographystyle{IEEEtran}
\bibliography{references}

\end{document}